\documentclass[a4paper]{article}

\usepackage{INTERSPEECH2016}

\usepackage{graphicx}
\usepackage{amssymb,amsmath,bm}
\usepackage{textcomp}

\usepackage{multirow}
\usepackage{xfrac}
\usepackage{array}
\usepackage{booktabs}
\usepackage{float}

\newcommand{\PreserveBackslash}[1]{\let\temp=\\#1\let\\=\temp}
\newcolumntype{C}[1]{>{\PreserveBackslash\centering}p{#1}}
\newcolumntype{R}[1]{>{\PreserveBackslash\raggedleft}p{#1}}
\newcolumntype{L}[1]{>{\PreserveBackslash\raggedright}p{#1}}

\ninept

\title{Learning Multiscale Features Directly From Waveforms}

\makeatletter
\def\name#1{\gdef\@name{#1\\}}
\makeatother \name{{\em Zhenyao Zhu* $^{1,2}$, Jesse H. Engel* $^{1}$, Awni Hannun$^1$}}

\address{$^1$Baidu Silicon Valley AI Lab (SVAIL) \\
  $^2$The Chinese University of Hong Kong \\
  \footnotesize{$^*$Authors contributed equally to this work} \\
  {\small \tt zhuzychn@gmail.com, jengel@baidu.com, awnihannun@baidu.com}
}

\begin{document}

\maketitle
\begin{abstract}

Deep learning has dramatically improved the performance of speech recognition systems through learning hierarchies of features optimized for the task at hand.  
However, true end-to-end learning, where features are learned directly from waveforms, has only recently reached the performance of hand-tailored representations based on the Fourier transform.
In this paper, we detail an approach to use convolutional filters to push past the inherent tradeoff of temporal and frequency resolution that exists for spectral representations. 
At increased computational cost, we show that increasing temporal resolution via reduced stride and increasing frequency resolution via additional filters delivers significant performance improvements.
Further, we find more efficient representations by simultaneously learning at multiple scales, leading to an overall decrease in word error rate on a difficult internal speech test set by 20.7\% relative to networks with the same number of parameters trained on spectrograms.

\end{abstract}
\noindent{\bf Index Terms}: Speech Recognition, Multiscale Learning, Convolutional Neural Networks, Raw Waveforms

\section{Introduction}
\label{sec:methodology}

Raw speech waveforms are densely sampled in time, and thus require downsampling to make many analysis techniques computationally tractable. 
For speech recognition, this presents the challenge to reduce the number of timesteps in the signal without throwing away relevant information.
Representations based on the Fourier transform have proven effective at this task as the transform forms a complete basis for signal reconstruction.
Deep learning's recent success in speech recognition is based on learning feature hierarchies atop these representations \cite{hinton2012speech, amodei2015deepspeech2}. 

There has been increasing focus on extending this end-to-end learning approach down to the level of the raw waveform. 
A popular approach is pass the waveform through strided convolutions, or networks connected to local temporal frames, often followed by a pooling step to create invariance to phase shifts and further downsample the signal \cite{palaz2015analysis, sainath2015learning, golik2015convolutional, tuske2014acoustic, jaitly2011rbm, dieleman2014end2end}. 
While some studies find inferior performance for convolutional filters learned in this way, deeper networks have recently matched the performance of hand-engineered features on large vocabulary speech recognition tasks \cite{sainath2015learning}.

Features based on the Fourier transform are computationally efficient, but exhibit an intrinsic tradeoff of temporal and frequency resolution. 
Convolutional filters decouple time and frequency resolution as the number of filters and stride are chosen independently. 
Despite this, a filter bank is constrained by its window size to a single scale.
Herein, we explore jointly learning filter banks at multiple different scales on raw waveforms. 
Multiscale convolutions have already been successfully applied to address tasks in the computer vision field, such as image classification \cite{szegedy2014going}, scene labeling \cite{farabet2013learning}, and gesture detection \cite{neverova2014multi}.
These successful applications exploit structure at different scales, which encourages us to explore multiscale representations for waveforms as well.

\begin{figure}[t]
    \includegraphics[width=230pt]{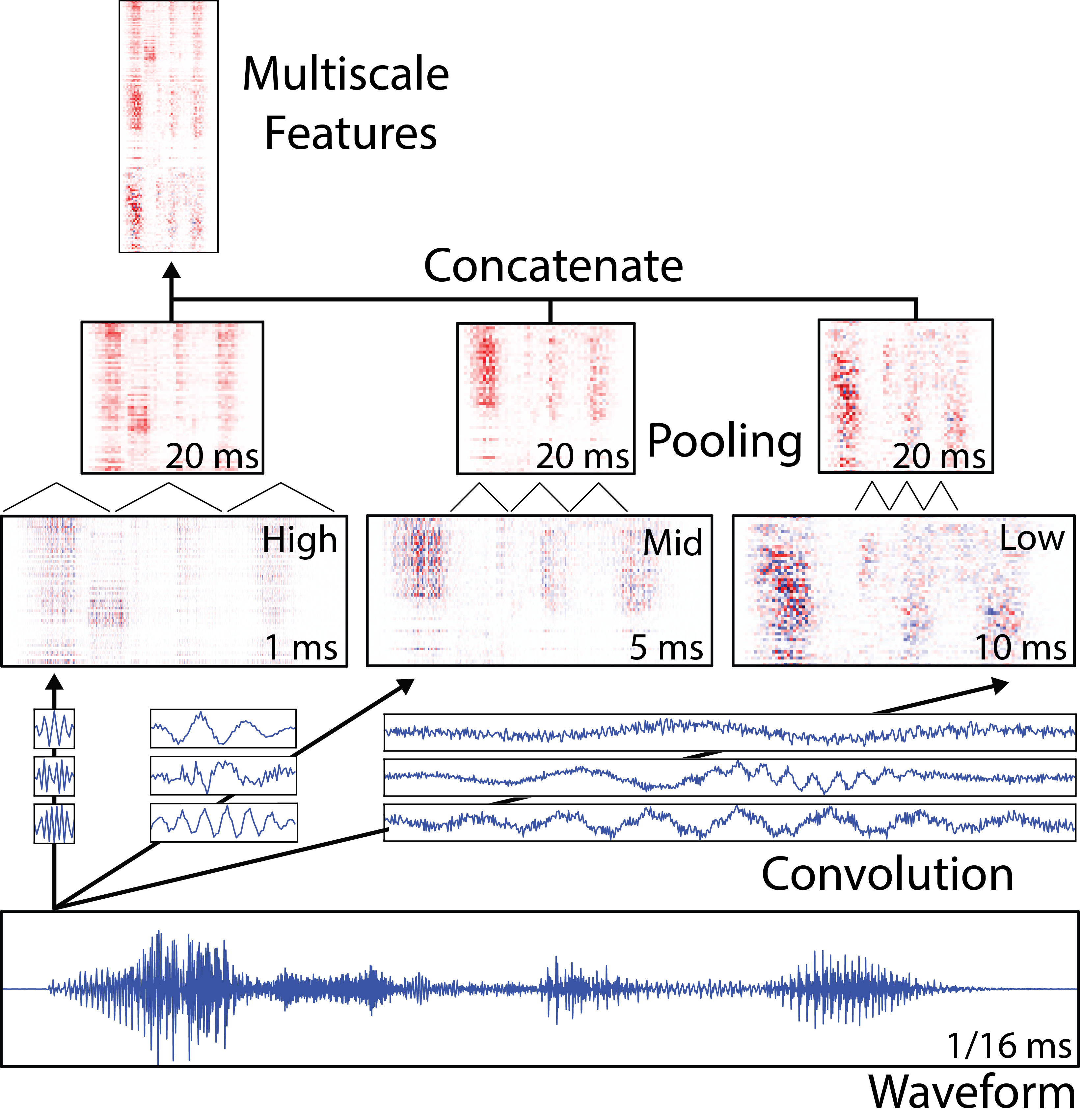}
    \caption{\footnotesize \it Diagram of multiscale convolutions for learning directly from waveforms. Sampled at 16kHz, the waveform is originally 1/16ms per frame. Three convolutions with different window sizes and strides are applied, leading to feature maps with High (1ms/frame), Mid (5ms/frame), and Low (10ms/frame) temporal resolution. Next, max-pooling and concatenation ensure a consistent sampling frequency (20ms/frame) for the rest of the network. The diagram shows real feature maps that are extracted by applying learned features (also shown) to a recording of one of the authors saying "I like cats" in Mandarin.}
    \label{fig:diagram}
\end{figure}

Further, multiscale convolutions enable us to split the spectrum into different filter banks with independent choice of stride, window size, and number of filters.
To learn high frequency features, filters with a short window are applied at a small stride on the waveforms. 
Low-frequency features, on the contrary, employ a long window that can be applied at a larger stride. 
Finally, based on the needs of the speech recognition system we can then independently tune the number of filters for the different frequency bands. 

While much research has already been conducted on learning directly from waveforms for speech recognition, the unique contributions of this paper are threefold:
\begin{itemize}
  \item We perform with an in-depth analysis of scaling to low strides and large numbers of filters and discover that a convolutional front end can significantly outperform Fourier features by independently tuning temporal and frequency resolution, at the cost of additional computation and memory.
  \item We propose a new multiscale convolutional front end, composed of concatenated filters with different window sizes, that requires less computation and outperforms features learned on a single scale (20.7\% relative to spectrogram baseline).
  \item We find that multiscale features naturally learn the frequencies they can most efficiently represent, with large and small windows learning low and high frequencies respectively. This contrasts with the single scale features which try to cover the entire frequency spectrum regardless of window size.
\end{itemize}

\section{Experimental Setup}
\label{sec:experiments}

\begin{figure}[t]
    \includegraphics[width=230pt]{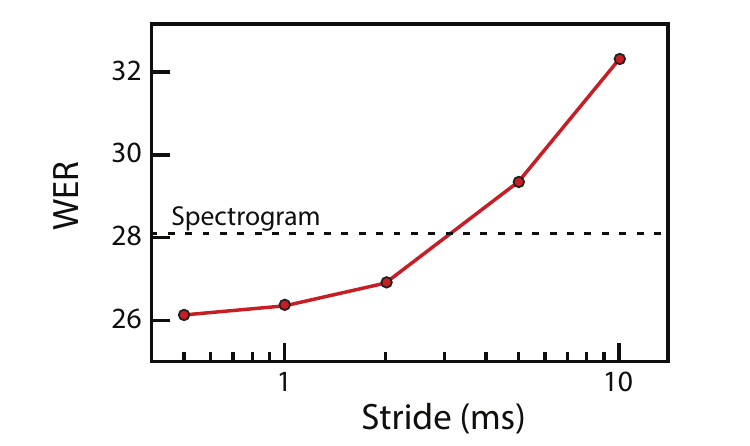}
    \caption{\footnotesize \it Stride matters. For a single scale of convolution, WER reduces with decreasing stride, passing the spectrogram baseline at 2ms stride. Smaller strides in convolution have larger strides in max-pooling, keeping the 20ms/frame context constant. While more computationally efficient, larger strides correspond to sparsely subsampling the data, and lose information compared with pooling.}
    \label{fig:wer_strides}
\end{figure}

\begin{figure*}[t]
    \includegraphics[width=\textwidth]{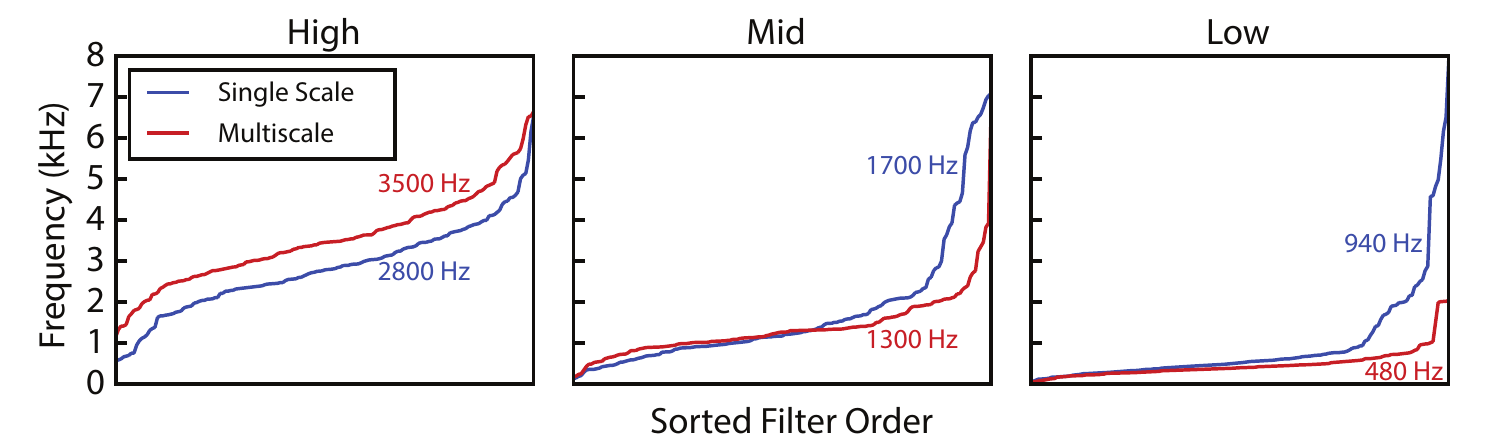}
    \caption{\footnotesize \it Multiscale features naturally separate according to frequency. Plots of spectral centroid of the learned filters (sorted by frequency) when scales are learned separately (blue) versus jointly (red). For quantitative comparison, the mean filter frequencies of each bank is also shown. In the separate case, each filter bank tries to span the entire frequency range, with larger windows having more emphasis on low frequencies. Learned jointly, the filter banks split responsibilities based on what they efficiently can represent, with smaller windows putting more emphasis on high frequencies and larger windows dedicating more filters to low frequencies.}
    \label{fig:MultiSpec}
\end{figure*}

The experimental design of this study is modelled after our previous work on end-to-end speech recognition \cite{hannun2014deepspeech, amodei2015deepspeech2}. 
However, to decrease the experimental latency, we train on a reduced version of the model and a subset of the training data.
The basic architecture is shown in Table~\ref{tab:Architecture}.
While we vary the front end processing, the backend remains the same: a convolutional (through time) layer, followed by three bidirectional simple recurrent layers, and a fully connected layer.
Batch normalization \cite{ioffe2015batchnorm}, is employed between each layer, but not between individual timesteps \cite{amodei2015deepspeech2}. 
Rectified linear unit (ReLU) activation functions are used for all layers, including between timesteps.
We use the Connectionist Temporal Classification (CTC) cost function to integrate over all possible alignments between the network outputs and characters of the English alphabet \cite{graves2006connectionist}.

Training is conducted on 2,400 hours of audio randomly sampled from 12,000 hours of data.
The training data is drawn from a diverse collection of sources including read, conversational, accented, and noisy speech \cite{amodei2015deepspeech2}.  
At each epoch, 40\% of the utterances are randomly selected to have background noise (superpositions of YouTube clips) added at signal-to-noise ratios ranging from 0dB to 15dB \cite{hannun2014deepspeech}.
All input data (either spectrogram or waveform) is sampled at 16kHz, and normalized so that each input feature has zero mean and unit variance.
For models that learn directly from {\it waveforms}, no other preprocessing is applied and the network inputs have 1 feature per 1/16ms frame.
For models that learn from {\it spectrograms}, linear FFT features are extracted with a hop size of 10ms, and window size of 20ms. 
The network inputs are thus spectral magnitude maps ranging from 0-8kHz with 161 features per 10ms frame.

\begin{table}[t]
    \vspace{10pt}    
    \centering
    \begin{tabular}{| c | c |} 
        \hline 
        Connectionist Temporal Classification (CTC) \\
        \hline
        Fully Connected (1770) \\
        \hline
        ... \\
        3x Bidirectional Recurrent (1770) \\
        ... \\
        \hline
        1-D Convolution (1770, window 11, stride 1) \\
        \hline
        Max-Pooling (20ms/frame) \\
        \hline
        Waveform Convolution or Spectrogram (161) \\
        \hline
        Waveform \\
        \hline
    \end{tabular}
    \caption{\footnotesize \it Neural network architectures explored in this study. The architecture is held constant except for the "Waveform Convolution" layer. The baseline (Spectrogram) is compared against different single and multiscale convolutions. }
    \label{tab:Architecture}
\end{table}

\begin{table}[t]
    \vspace{5pt}
    
    \centering
    \footnotesize
    \begin{tabular}{| c | c | c | c | c || c | } 
    \hline 
    \multirow{2}{*}{Type} & \multicolumn{3}{c|}{Spectrogram~/~Convolution}& Pooling & \multirow{2}{*}{WER(\%)} \\
    \cline{2-5}
     & \# Features & Window & Stride & Stride & \\
    \hline
    FFT & 161 & 20ms & 10ms & 2 & \textbf{28.10}\\
    \hline
    wav & 161 & 20ms & 10ms & 2 & 32.31\\
    wav & 161 & 20ms & 5ms & 4 & 29.35\\
    wav & 161 & 20ms & 2ms & 10 & 26.90\\
    wav & 161 & 20ms & 1ms & 20 & 26.35\\
    wav & 161 & 20ms & 0.5ms & 40 & \textbf{26.13}\\
    \hline 
    \end{tabular}
    \caption{\footnotesize \it Single scale waveform convolution outperforms the spectrogram baseline at low strides. The trend is visualized in Figure~\ref{fig:wer_strides}.}
    \label{tab:wer_strides}
\end{table}

We train using stochastic gradient descent with Nesterov momentum and a batch size of 128. 
Hyperparameters are tuned for each model by optimizing a hold-out set.
Typical values are a learning rate of 3e-4 and momentum of 0.99, and training converges after 20 epochs.
Following \cite{amodei2015deepspeech2}, we sort the first epoch by utterance length (SortaGrad), to promote stability of training on long utterances. 
While the CTC-trained acoustic model learns a rudimentary language model itself from the training data, for testing, we supplement it with a Kneser-Ney smoothed 5-gram model that is trained using the KenLM toolkit \cite{heafield2013kenlm} on cleaned text from the Common Crawl Repository.
Decoding is done via beam search, where a weighted combination of the acoustic model and language model with an added word insert penalty is used as the value function.
Test set word error rates (WER) are reported on a difficult in-house test set of 2048 utterances, diversely composed of noisy, conversational, voice-command, and accented speech. 
The test set is collected internally and from industry partners and is not represented in the training data.

As previously observed \cite{amodei2015deepspeech2}, deep neural networks trained on sufficient data perform better as the model size grows. 
In order to make fair comparisons, the number of parameters of the models used in our experiments is held constant at 35M. 
We are aware that the results are not directly comparable to literature due to the use of proprietary datasets.
However, we attempt to tightly control our experiments rather than focus on overall performance, with the aim that the conclusions will generalize to other architectures and datasets as well.
If optimizing for performance, it is worth noting that the WER drops by $\sim$50\% when training on all 12,000 hours of data with 7 bidirectional layers (70M parameters) in the backend.

\section{Results}
\label{sec:results}

\subsection{Decoupling temporal resolution}
\label{sec:wav}

While many studies have compared convolution on raw waveforms to features such as spectrograms, MFCCs, and melscale filterbanks, they often compare the two at a similar strides and window sizes \cite{golik2015convolutional, tuske2014acoustic}. 
Spectrograms can employ a high stride because of the unique analytic structure of the basis functions. 
Integrating twice, once over the real cosine and once over its imaginary sine counterpart, identifies both the phase and magnitude of the response. 
This is in many ways similar to performing a convolution over \emph{every timestep} and max-pooling, with the phase represented by the index at which the max occurs, but is much more computationally efficient to perform with an FFT. 

We decided to test this hypothesis that a convolution with low stride and pooling can find at least as good a basis as the spectrogram. 
Figure~\ref{fig:wer_strides} and Table~\ref{tab:wer_strides} show the effects of replacing the spectrogram with a single convolution and pooling layer of the same number of filters and window size. 
For each decrease in stride of the convolution, the stride of the pooling is increased to give a consistent total stride of 20ms, which is the same as the stride of the spectrogram with pooling.  
At comparable strides to the spectrogram (10ms) the convolution is unable to perform as well, likely due to needing to represent phase shifts as well as frequency variation. 
However, as the stride dips below 2ms, the convolution asymptotically reaches a superior level of performance to the spectrogram. 
To be fair, this improved performance comes at increased computational cost and memory usage. 

\subsection{Multiscale features are more efficient}

Representing high frequency information with a large filter is difficult because of the many places the information can occur in the filter window.
Similarly, representing low frequency information with a small filter is challenging because separate filters are required for separate parts of the wave.
We hypothesized that applying convolution simultaneously at several scales could allow each scale to learn filters selective to the frequencies that it can most efficiently represent.
To test our hypothesis, we compare the performance of convolutional front ends with a constant number of filters at three different scales (High: (1ms window, 1/4ms stride), Mid: (4ms, 1ms), and Low: (40ms, 10ms)), to a front end employing all three scales (Diagramed in Figure~\ref{fig:diagram}.

Table~\ref{tab:wav_multi} shows that WER significantly decreases from High to Mid to Low frequency filter banks.
The improved performance of the lower frequency banks is evidence for the importance of low frequency vocalization features in speech recognition \cite{sainath2015learning}. 
Figure~\ref{fig:MultiSpec} displays the spectral centroids of each filter bank sorted by frequency, with the mean value printed alongside. 
It is clear that the High (2800Hz), Mid (1700Hz), and Low (940Hz) filter banks live up to their names, each learning filters that capture different frequency bands.
When we jointly learn several scales, the filters exhibit a an heightened preference for representing different bands.
Relaxing the requirement of each bank to cover the entire spectrum causes the High (3500Hz) banks to increase in mean frequency, and the Mid (1300Hz) and Low (480Hz) banks to decrease.
Further, the WER is lowest for the multiscale features, despite the fact that it has fewer parameters and three times less large filters.

\begin{figure*}[t]
    \includegraphics[width=\textwidth]{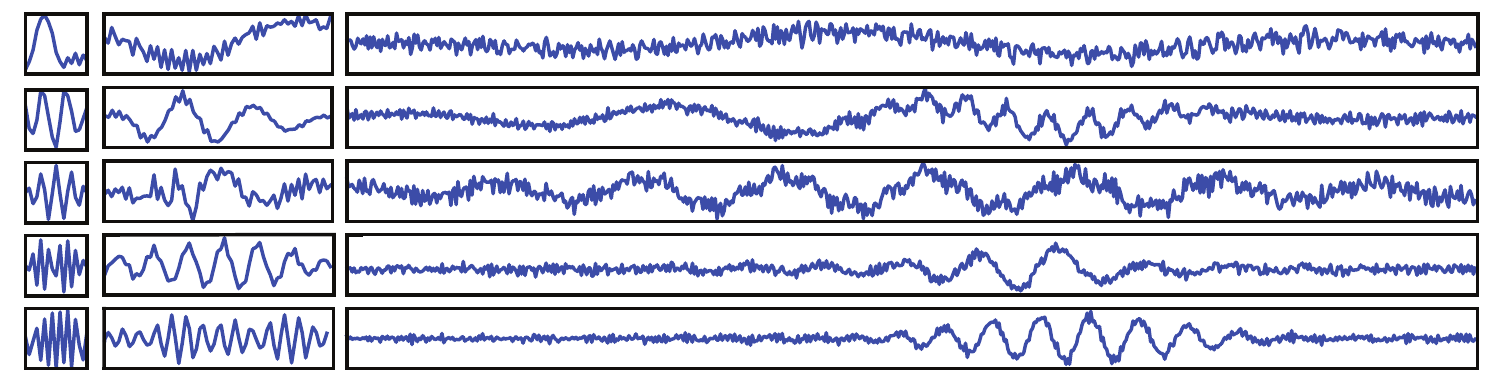}
    \caption{\footnotesize \it Representative learned multiscale filters. There is a clear preference for Fourier-like and wavelet representations, with varying degrees of high frequency noise. Some filters also show a combination of frequencies and localized transient structure. Phase shifted filter pairs are also found, suggesting the importance of phase information in speech recognition tasks.}
    \label{fig:filters}
\end{figure*}

\begin{table}[t]
    \vspace{5pt}    
    \centering
    \footnotesize
    \begin{tabular}{|m{42pt} | m{40pt} | m{40pt}|| m{25pt} |} 
        \hline 
        \multicolumn{3}{|c||}{\# Features} & \multirow{2}{*}{WER(\%)} \\
        \cline{1-3}
        High (1ms) & Mid (4ms) & Low (40ms) & \\
        \hline
        161 & 0 & 0 & 32.84  \\
        0 & 161 & 0 &  27.69 \\
        0 & 0 & 161 &  26.54 \\
        \hline
        61 & 50 & 50  & \textbf{25.67} \\
        \hline 
    \end{tabular}
    \caption{\footnotesize \it Multiscale vs. single scale features. Holding the number of filters constant, single scale convolution with High frequency (1ms window, 1/4ms stride), Mid frequency (4ms, 1ms) or Low frequency (40ms, 10ms), are outperformed by a combination of all three (as diagrammed in Figure~\ref{fig:diagram}).}
    \label{tab:wav_multi}
\end{table}

\subsection{Decoupling frequency resolution}

With methods based on the Fourier transform, there is an intrinsic tradeoff of temporal and frequency resolution.
As the number of basis functions increases, helping identify which frequencies are present, so does the window size, which smears knowledge of when they occur.
Decreasing the stride cannot increase temporal resolution, it can only provide more samples of the smeared signal.
Convolutions do not suffer from this same tradeoff, as the temporal resolution is limited by the number of filters and temporal resolution by the stride, both of which are independent.
This advantage comes at the added cost of increased computation and memory, both by increasing number of filters or decreasing stride.

We explore the value of increased resolution by performing the same multiscale experiments as the previous section, but increasing the number of filters. 
Table~\ref{tab:many_filters} demonstrates that increasing the number of filters, even by a factor of 3, leads to a significant (8\% relative) improvement in WER. 
A fully-connected layer with output dimension of 161 is added above the concatenated feature maps in order to produce the same number of features as the previous experiments. 
Since using more filters at short strides is more costly in terms of both computation and memory, we specifically increase the number of filters at long strides. 
Further increasing the number of filters, and expanding the size of the bottle neck leads to smaller gains. showing an impressive 20.7\% cumulative improvement in WER relative to the spectrogram baseline. 

\begin{table}[t]
    \vspace{5pt}    
    \centering
    \footnotesize
    \begin{tabular}{|m{42pt} | m{40pt} | m{45pt}|| m{25pt} |} 
        \hline 
        \multicolumn{3}{|c||}{\# Features} & \multirow{2}{*}{WER(\%)} \\
        \cline{1-3}
        High (1ms) & Mid (4ms) & Low (40ms) & \\
        \hline
        61 & 50 & 50 & 25.67  \\
        161 & 161 & 161 &  23.78 \\
        160 & 320 & 640 &  23.52 \\
        160 & 320 & 640 &  \textbf{23.28$^{*}$} \\
        \hline 
    \end{tabular}
    \caption{\footnotesize \it Decoupling time and frequency resolution. Window size shown in parentheses. Adding additional filters at the same stride (1/4ms, 1ms, 10ms) significantly improves performance. All models have an additional bottleneck layer (Fully Connected, 161) inserted after pooling to maintain the same number of features as previous experiments. $^{*}$Increasing the bottleneck layer from 161 to 800 leads to a small improvement.}
    \label{tab:many_filters}
\end{table}

\section{Discussion}
\label{sec:Discussion}

In this paper, we have consistently demonstrated that learning features directly from waveforms can outperform spectrograms, especially when applied at multiple scales.
However, several interesting research questions remain to be answered before such techniques likely see widespread adoption. 

Learning convolutional filters overcomes the time/frequency tradeoff of Fourier representations, but with considerable computational and memory cost. 
Many modern state of the art systems train on clusters of GPUs, where memory is precious, as requiring memory transfer between GPU and CPU can be prohibitively slow for training. 
This is especially problematic for training on long utterances, where the amount of memory required to save all the activations increases both with the number of filters and the reduction of stride.
It remains to be seen whether the power of learned input features can be combined with the efficiency of analytic signal transformations such as the Fourier transform.
One such approach could be to learn basis functions in the real and imaginary domain by performing backpropagation through the Hilbert transformation, enabling the use of larger strides.
Alternatively, learned features can be fixed and used to augment other fixed features at train time.
Sainath et al. \cite{sainath2015learning} found noticeable improvements from supplementing log-mel filterbanks in such a manner. 

While learned features outperformed spectrograms feeding into temporal convolution in this study, many state of the art systems apply two-dimensional convolutions to their inputs \cite{amodei2015deepspeech2, sercu2015conv}. 
Our learned features underperform in this context, which is understandable as they are not spectrally ordered, and lack spatial structure. 
Regularization techniques such as \cite{ranzato2007invariant} could perhaps be key to learning ordered filter maps with useful structure. 

In our experiments, we made sure to downsample each scale equally with appropriate stride such that the signals can be concatenated for the later recurrent layers.
This temporal pooling only takes into account local structure and has no explicit knowledge of what information to preserve based on long-range dependencies.
Recently proposed architectures that operate simultaneously at different timescales, such as the Clockwork RNN \cite{koutnik2014clockwork}, could provide a more elegant way of combining multiscale signals. 
Beyond incorporating recurrence, low frequency features that require fewer temporal samples could then also require less recurrent computation and facilitate modeling long-range structure. 

Finally, from observing representative filters learned at each scale in Figure~\ref{fig:filters}, we can see that there is some redundancy in the representation.
Some filter shapes appear similar at multiple scales. 
An interesting future direction could be to investigate learning features in a scale-free basis, similar to wavelets, where a reduced basis set could be applied across a range of scales. 

\newpage
\eightpt
\bibliographystyle{IEEEtran}

\bibliography{multiscale_conv}

\end{document}